\documentclass{article} 
\usepackage{nips13submit_e,times}
\usepackage{hyperref}
\usepackage{url}

\usepackage{natbib}
\usepackage{amsmath}
\usepackage{amssymb}
\usepackage{todonotes}

\newcommand{\expc}[1]{\text{E}[#1]}
\newcommand{\vari}[1]{\text{V}[#1]}
\newcommand{\abovespace}{}{}
\newcommand{\belowspace}{}{}
\newcommand{\Win}[0]{W_{\text{in}}}
\newcommand{\Wout}[0]{W_{\text{out}}}
\newcommand{\Wrec}[0]{W_{\text{rec}}}

\title{On Fast Dropout and its Applicability to Recurrent Networks}

\author{
Justin~Bayer, Christian~Osendorfer, Daniela~Korhammer, \\
\textbf{Nutan~Chen, Sebastian~Urban and Patrick~van~der~Smagt} \\
Lehrstuhl f\"ur Robotik und Echtzeitsysteme\\
Fakult\"at f\"ur Informatik\\
Technische Universit\"at M\"unchen \\
\texttt{bayer.justin@googlemail.com, osendorf@in.tum.de, korhammd@in.tum.de,} \\
\texttt{ntchen86@gmail.com, surban@tum.de, smagt@brml.org} \\
}

\nipsfinalcopy 

\begin{document}

\maketitle

\begin{abstract}
Recurrent Neural Networks (RNNs) are rich models for the processing of sequential data.
Recent work on advancing the state of the art has been focused on the optimization or modelling of RNNs, mostly motivated by adressing the problems of the vanishing and exploding gradients.
The control of overfitting has seen considerably less attention.
This paper contributes to that by analyzing fast dropout, a recent regularization method for generalized linear models and neural networks from a back-propagation inspired perspective.
We show that fast dropout implements a quadratic form of an adaptive, per-parameter regularizer, which rewards large weights in the light of underfitting, penalizes them for overconfident predictions and vanishes at minima of an unregularized training loss.
The derivatives of that regularizer are exclusively based on the training error signal.
One consequence of this is the absence of a global weight attractor, which is particularly appealing for RNNs, since the dynamics are not biased towards a certain regime.
We positively test the hypothesis that this improves the performance of RNNs on four musical data sets.

\end{abstract}

\section{Introduction}
Recurrent Neural Networks are among the most powerful models for sequential data.
The capabilty of representing any measurable sequence to sequence mapping to arbitrary accuracy \citep{hammer2000approximation} makes them universal approximators.
Nevertheless they were given only little attention in the last two decades due to the problems of vanishing and exploding gradients \citep{hochreiter1991untersuchungen,bengio1994learning,pascanu2012difficulty}.
Error signals either blowing up or decaying exponentially for events many time steps apart rendered them largely impractical for the exact problems they were supposed to solve.
This made successful training impossible on many tasks up until recently without resorting to special architectures or abandoning gradient-based optimization.
Successful application on tasks with long-range dependencies has thus relied on one of those two paradigms.
The former ist to  make use of long short-term memory (LSTM) \citep{hochreiter1997long}.
These approaches are among the best methods for the modelling of speech and handwriting \citep{graves2013speech, graves2008unconstrained, graves2013generating}.
The latter is to to rely on sensible initializations leading to echo-state networks~\citep{jaeger2003adaptive}.

The publication of \citep{martens2011learning} can nowadays be considered a landmark, since it was shown that even standard RNNs can be trained with the right optimization method.
While a sophisticated Hessian-free optimizer was employed initially, further research \citep{sutskever2013importance,bengio2012advances} has shown that carefully designed first-order methods can find optima of similar quality.

After all, the problem of underfitting standard RNNs can be dealt with to the extent that RNNs are practical in many areas, e.g., language modelling \citep{sutskever2011generating,mikolov2010recurrent}.
In contrast, the problem of overfitting in standard RNNs has (due to the lack of necessity) been tackled only by few.
As noted in \citep{pascanu2012difficulty}, using priors with a single optima on the parameters may have detrimental effects on the representation capability of RNNs: a global attractor is constructed in parameter space.
In the case of a prior with a mode at zero (e.g. an $L2$ regularizer) this biases the network towards solutions which lets information die out exponentially fast in time, making it impossible to memorize events for an indefinite amount of time.

\cite{graves2011practical} proposes to stochastically and adaptively distort the weights of LSTM-based RNNs, which is justified from the perspective of variational Bayes and the minimum description length principle.
Overfitting is practically non-existent in the experiments conducted.
It is untested whether this approach works well for standard RNNs--along the lines of the observations of \cite{pachitariu2013regularization} one might hypothesize that the injected noise disturbs the dynamics of RNNs too much and leads to divergence during training.

The deep neural network community has recently embraced a regularization method called dropout \citep{hinton2012improving}.
The gist is to randomly discard units from the network during training, leading to less interdependent feature detectors in the intermediary layers.
Here, dropping out merely means to set the output of that unit to zero. 
An equivalent view is to set the complete outgoing weight vector to zero of which it is questionable whether a straight transfer of dropout to RNNs is possible.
The resulting changes to the dynamics of an RNN during every forward pass are quite dramatic.
This is the reason why \cite{pachitariu2013regularization} only use dropout on those parts of the RNN which are not dynamic., i.e.~the connections feeding from the hidden into the output layer.

Our contribution is to show that using a recent smooth approximation to dropout \citep{wang2013fast} regularizes RNNs effectively.
Since the approximation is deterministic, we may assert that all dynamic parts of the network operate in reasonable regimes.
We show that fast dropout does not keep RNNs from reaching rich dynamics during training, which is not obvious due to the relation of classic dropout to L2 regularization \citep{wager2013dropout}.

The structure of the paper is as follows.
We will first review RNNs and fast dropout (FD) \citep{wang2013fast}.
A novel analysis of the derivatives of fast dropout leads to an interpretation where we can perform a decomposition into a loss based on the average output of a network's units and a regularizer based on its variance. 
We will discuss why this is a form that is well suited to RNNs and consequently conduct experiments that confirm our hypothesis.

\section{Methods}
In this section we will first review RNNs and fast dropout.
We will then introduce a novel interpretation of what the fast dropout loss constitutes in section \ref{subsubsec-bp} and show relationships to two other regularizers.
\subsection{Recurrent Neural Networks}
\label{sub-sec:rnn}
\paragraph{}
We will define RNNs in terms of two components.
For one, we are ultimately interested in an output $\mathbf{y}$, which we can calculate given the parameters $\theta$ of a network and some input $\mathbf{x}$.
Secondly, we want to learn the parameters, which is done by the design and optimization of a function of the parameters $\mathcal{L}(\theta)$, commonly dubbed loss, cost, or error function.

\paragraph{Calculating the Output of an RNN}
Given an input sequence $\mathbf{x} = (x_1, \dots, x_T), x_t \in \mathbb{R}^\kappa$ we produce an output $\mathbf{y} = (y_1, \dots, y_T), y_t \in \mathbb{R}^\omega$ which is done via an intermediary representation called the hidden state layer $\mathbf{h} = (h_1, \dots, h_T), h_t \in \mathbb{R}^\gamma$. $\kappa$, $\omega$, and $\gamma$ are the dimensionalities of the inputs, outputs, and hidden state at each time step.
Each component of the layers is sometimes referred to as a unit or a ``neuron''.
Depending on the associated layer, these are then input, hidden, or output units.
We will also denote the set of units which feed into some unit $i$ as the incoming units of $i$.
The units into which a unit $i$ feeds are called the outgoing units of $i$.
For a recurrent network with a single hidden layer, this is done via iteration of the following equations from $t=1$ to $T$:
\begin{eqnarray*}
    h_t &=& f_h(x_t \Win + h_{t-1} \Wrec + b_h), \\
    y_t &=& f_y(h_t \Wout + b_y),
\end{eqnarray*}
where $\{\Win, \Wout, \Wrec\}$ are weight matrices and $\{b_h, b_y\}$ bias vectors.
These form the set of parameters $\theta$ together with initial hidden state $h_0$.
The dimensionalities of all weight matrices, bias vectors, and initial hidden states are determined by the dimensionalities of the input sequences as well as desired hidden layer and output layer sizes.
The functions $f_h$ and $f_y$ are so-called transfer functions and mostly coordinate-wise applied nonlinearities.
We will call the activations of units \emph{pre-synaptic} before the application of $f$ and \emph{post-synaptic} afterwards.
Typical choices include the \emph{logistic sigmoid} $f(\xi) = {{1} \over {1 + \exp(-\xi)}}$, tangent hyperbolicus and, more recently, the rectified linear $f(\xi) = \max(\xi, 0)$ \citep{zeiler2013rectified}.
If we set the recurrent weight matrix $\Wrec$ to zero, we recover a standard neural network \cite{bishop1995neural} applied independently to each time step.


\paragraph{Loss Function and Adaption of Parameters}
We will restrict ourselves to RNNs for the supervised case, where we are given a data set $\mathcal{D} = \{(\mathbf{x}_i, \mathbf{z}_i)\}_{i=1}^N$ consisting of $N$ pairs with $\mathbf{x}_i \in \mathbb{R}^{T \times \kappa}$ and $\mathbf{z}_i \in \mathbb{R}^{T \times \omega}$. 
Here $T$ refers to the sequence length, which we assume to be constant over the data set.
We are interested to adapt the parameters of the network $\theta$ in a way to let each of its outputs $\mathbf{y}_i \in \mathbb{R}^{T \times O}$ be close to $\mathbf{z}_i$.
Closeness is typically formulated as a loss function, e.g. the mean squared error $\mathcal{L}_{\text{mse}}(\theta) = \sum_i ||\mathbf{z}^{(i)} - \mathbf{y}^{(i)}||_2^2$ or the binary cross entropy $\mathcal{L}_{\text{bce}}(\theta) = \sum_{i} z^{(i)} \log y^{(i)} + (1 - z^{(i)}) \log (1 - y^{(i)})$.
If a loss $\mathcal{L}$ is locally differentiable, finding good parameters can be performed by gradient-based optimization, such as nonlinear conjugate gradients or stochastic gradient descent.
The gradients can be calculated efficiently via back-propagation through time (BPTT) \citep{rumelhart1986learning}.

\subsection{Fast Dropout}
\label{sub-sec:fd}
In fast dropout \citep{wang2013fast}, each unit in the network is assumed to be a random variable.
To assure tractability, only the first and second moments of those random variables are kept, which suffices for a very good approximation.
Since the pre-synaptic activation of each unit is a weighted sum of its incoming units (of which each is dropped out with a certain probability) we can safely assume Gaussianity for those inputs due to the central limit theorem.
As we will see, this is sufficient to find efficient ways to propagate the mean and variance through the non-linearity $f$.

\subsubsection{Forward propagation}
We will now inspect the forward propagation for a layer into a single unit, that is 
\begin{eqnarray*}
a =& (\mathbf{d} \circ \mathbf{x})^T \mathbf{w} \\
y =& f(a),
\end{eqnarray*}
where $\circ$ denotes the element-wise product and $f$ is a non-linear transfer function  as before.
Let the input layer $\mathbf{x}$ to the unit be Gaussian distributed with diagonal covariance by assumption: $\mathbf{x} \sim \mathcal{N}(\mu_\mathbf{x}, \sigma^2_\mathbf{x})$.
Furthermore, we have Bernoulli distributed variables indicating whether an incoming unit is not being dropped out organized in a vector $\mathbf{d}$ with $d_i \sim \mathcal{B}(p)$, $p$ being the complementary drop out rate.
The weight vector $\mathbf{w}$ is assumed to be constant.

A neural network will in practice consist of many such nodes, with some of them, the output units, directly contributing to the loss function $\mathcal{L}$.
Others, the input units, will not stem from calculation but come from the data set.
Each component of $\mathbf{x}$ represents an incoming unit, which might be an external input to the network or a hidden unit.
In general, $y$ will have a complex distribution depending highly on the nature of $f$.
Given that the input to a function $f$ is Gaussian distributed, we obtain the mean and variance of the output as follows:
\begin{eqnarray*}
\expc{y} = f_{\mu}(a) &=& \int f(x) \mathcal{N}(x|\expc{a}, \vari{a}) dx, \\
\vari{y} = f_{\sigma}(a) &=& \int (f(x) - f_{\mu}(a))^2 \mathcal{N}(x|\expc{a}, \vari{a}) dx.
\end{eqnarray*}
Forward propagation through the non-linearity $f$ for calculation of the post-synaptic activation can be approximated very well in the case of the \emph{logistic sigmoid} and the \emph{tangent hyperbolicus} and done exactly in case of the \emph{rectifier} (for details, see \cite{wang2013fast}).
While the rectifier has been previously reported to be a useful ingredient in RNNs \citep{bengio2012advances} we found that it leads to unstable learning behaviour in preliminary experiments and thus neglected it in this study, solely focusing on the tangent hyperbolicus.
Other popular transfer functions, such as the softmax, need to be approximated either via sampling or an unscented transform \citep{julier1997new}.

To obtain a Gaussian approximation for $a = (\mathbf{d} \circ \mathbf{x})^T \mathbf{w}$, we will use $\hat{a} \sim \mathcal{N}(\expc{a}, \vari{a})$.
The mean and variance of $a$ can be obtained as follows. 
Since $\mathbf{d}$ and $\mathbf{x}$ are independent it follows that
\begin{eqnarray}
\expc{a} = \expc{(\mathbf{x}\circ\mathbf{d})^T\mathbf{w}} = (\expc{\mathbf{x}} \circ \expc{\mathbf{d}})^T\mathbf{w}.
\label{eq:fd-exp}
\end{eqnarray}
For independent random variables $A$ and $B$, $\vari{AB} = \vari{A}\expc{B}^2+\expc{A}^2\vari{B} + \vari{A}\vari{B}$.
If we assume the components of $\mathbf{x}$ to be independent, we can write \footnote{In contrast to \cite{wang2013fast}, we do not drop the variance not related to dropout.  In their case, $\vari{a} = p(1-p)\mu_{\mathbf{x}}^{2T}\mathbf{w}^2$ was used instead of Equation (\ref{eq:fd-var}).}
\begin{eqnarray}
\vari{a}
= (p(1-p)\mu_{\mathbf{x}}^2 + p\sigma_{\mathbf{x}}^2)^T\mathbf{w}^2.
\label{eq:fd-var}
\end{eqnarray}
Furthermore the independency assumption is necessary such that the Lyapunov condition is satisfied \citep{lehmann1999elements} for the the central limit theorem to hold,  ensuring that $a$ is approximately Gaussian.

Propagating the mean and the variance through $f$ via $f_{\mu}$ and $f_{\sigma}$ suffices for determining the pre-synaptic moments of the outgoing units.
At the output $\mathbf{y}$ of the whole model, we will simplify matters and ignore the variance.
Some loss functions take the variance into account (e.g., a Gaussian log-likelihood as done in \citep{bayer2013training}).
Sampling can be a viable alternative as well.

\paragraph{Fast Dropout for RNNs} The extension of fast dropout to recurrent networks is straightforward from a technical perspective.
First, we note that we can concatenate the input vector at time step $t$, $x_t$ and the hidden state at the previous layer $h_{t-1}$ into a single vector: $c_t = [h_{t-1}, x_t]$.
We obtain a corresponding weight matrix by concatenation of the input to hidden and recurrent weight matrices $\Win$ and $\Wrec$: $W_c = [\Wrec, \Win]$.
We can thus reduce the computation to the step from above.

%

\subsubsection{Beyond the Backward Pass: A Regularization Term}
\label{subsubsec-bp}
Given the forward pass, we used automatic differentiation with Theano \citep{bergstra2010theano} to calculate the gradients.
Nevertheless, we will contribute a close inspection of the derivatives.  This will prove useful since it makes it possible to interpret fast dropout as an additional regularization term \emph{independent of the exact choice of loss function}.

Consider a loss $\mathcal{L}(\mathcal{D}; \theta)$ which is a function of the data $\mathcal{D}$ and parameters $\theta$ \footnote{We will frequently omit the explicit dependency on $\mathcal{D}$ and $\theta$ where clear from context.}.
In machine learning, we wish this loss to be minimal under unseen data $\mathcal{D}$ although we only have access to a training set $\mathcal{D}_{\text{train}}$.
A typical approach is to optimize another loss $\mathcal{J}(\mathcal{\mathcal{D}_{\text{train}}}; \theta)$ as a proxy in the hope that a good minimum of it will correspond to a good minimum of $\mathcal{L}$ for unseen data.
Learning is often done by the optimization of $\mathcal{J} = \mathcal{L} + \mathcal{R}$, where $\mathcal{R}$ is called a regularizer.
A common example of a regularizer is to place a prior on the parameters, in which case it is a function of $\theta$ and corresponds to the log-likelihood of the parameters.
For weight decay, this is a spherical Gaussian with inverse scale $\lambda$, i.e. $\mathcal{R}_{\text{wd}}(\theta) = \lambda ||\theta||_2^2$.
Regularizers can be more sophisticated, e.g. \cite{rifai2011manifold} determine directions in input space to which a model's outputs should be invariant.
More recently, dropout (i.e. non-fast dropout) for generalized linear models has been intepreted as a semi-supervised regularization term encouraging confident predictions by \cite{wager2013dropout}.

While it seems difficult to bring the objective function $\mathcal{J}_{\text{fd}}(\mathcal{D};\theta)$ of fast dropout into the form of $\mathcal{L} + \mathcal{R}$, it is possible with the derivatives of each node.
For this, we perform back-propagation like calculations.

Let $a = (\mathbf{d} \circ \mathbf{x})^T\mathbf{w}$ and $y = f(a)$ be the pre- and post-synaptic activations of a component of a layer in the network.
First note that ${\partial \mathcal{J} / \partial w_i} = {\partial \mathcal{J} / \partial a} \cdot {\partial a / \partial w_i}$ according to the chain rule.
Since $a$ is a random variable, it will be described in one of two forms.
In the case of a Gaussian approximation, we will summarize it in terms of its mean and variance; this approach is used if propagation through $f$ is possible in closed form.
In the case of sampling, we will have a single instantiation $\hat{a}$ of the random variable, which we can propagate through $f$.
An analysis of both cases is as follows.

\paragraph{Gaussian approximation} We find the derivative of $\mathcal{J}$ with respect to one of its incoming weights $w_i$ to be
\begin{eqnarray*}
{\partial \mathcal{J} \over \partial w_i} =
    {\partial \mathcal{J} \over \partial \expc{a}} {\partial \expc{a} \over \partial w_i}
     + {\partial \mathcal{J} \over \partial \vari{a}} {\partial \vari{a} \over \partial w_i}.
\end{eqnarray*}
We know that $\expc{a} = (\mathbf{x} \circ \mathbf{d})^T \mathbf{w}$ and thus $\partial \expc{a} / \partial w_i = x_i d_i$.
This can be recognized as the standard back-propagation term if we consider the dropout variable $d_i$ as fixed.
We will thus define
\begin{eqnarray}
{\partial \mathcal{L}^a \over \partial w_i} :=
    {\partial \mathcal{J} \over \partial \expc{a}} {\partial \expc{a} \over \partial w_i},
\label{eq:fd-l}
\end{eqnarray}
and subsequently refer to it as the local derivative of the training loss.
The second term can be analysed similarly.
We apply the chain-rule once more which yields
\begin{eqnarray*}
    {\partial \mathcal{J} \over \partial \vari{a}} {\partial \vari{a} \over \partial w_i} = 
     \underbrace{\partial \mathcal{J} \over \partial \vari{a}}_{:= \delta^a} {\partial \vari{a} \over \partial w_i^2} {\partial w_i^2 \over \partial w_i}.
\end{eqnarray*}
for which any further simplification of $\delta^a$ depends on the exact form of $\mathcal{J}$.
The remaining two factors can be written down explicitly, i.e.
\begin{eqnarray*}
    {\partial \vari{a} \over \partial w_i^2} &=& p(1-p)\expc{x_i}^2 + p \vari{x_i}, \\
    {\partial w_i^2 \over \partial w_i} &=& 2w_i.
\end{eqnarray*}
Setting 
\begin{eqnarray*}
    \eta^a_i :=& |\delta^a| {\partial \vari{a} \over \partial w_i^2}\\
    =& |\delta^a| p \big [(1-p)\expc{x_i}^2 + \vari{x_i} \big] \\
    >& 0
\end{eqnarray*}
we conclude that 
\begin{eqnarray*}
    {\partial \mathcal{J} \over \partial \vari{a}} {\partial \vari{a} \over \partial w_i} &=& 2~\text{sgn}(\delta^a)\eta^a w_i \\
    &=:& {\partial \mathcal{R}_{\text{approx}}^a \over \partial w_i}.
\end{eqnarray*}
In alignment with Equation (\ref{eq:fd-l}) this lets us arrive at 
\begin{eqnarray*}
{\partial \mathcal{J} \over \partial w_i} =
    {\partial \mathcal{L}^a \over \partial w_i}
    + {\partial \mathcal{R}_{\text{approx}}^a \over \partial w_i},
\end{eqnarray*}
and offers an interpretation of fast dropout as an additive regularization term.
An important and limiting aspect of this decomposition is that it only holds locally at $a$.

We note that depending on the sign of the error signal $\delta^a$, fast dropout can take on three different behaviours:
\begin{itemize}
\item $\delta^a = 0$ The error signal is zero and thus the variance of the unit considered to be optimal for the loss. The fast dropout term vanishes; this is especially true at optima of the overall loss.
\item $\delta^a < 0$ The unit should increase its variance. The exact interpretation of this depends on the loss, but in many cases this is related to the expectation of the unit being quite erroneous and leads to an increase of scatter of the output. The fast dropout term encourages a quadratic growth of the weights.
\item $\delta^a > 0$ The unit should decrease its variance. As before, this depends on the exact loss function but will mostly be related to the expectation of the unit being quite right which makes a reduction of scatter desirable. The fast dropout term encourages a quadratic shrinkage of the weights.
\end{itemize}

This behaviour can be illustrated for output units by numerically inspecting the values and gradients of the pre-synaptic moments given a loss.
For that we consider a single unit $y = f(a)$ and a loss $d(y, z)$ measuring the divergence of its output to a target value $z$.
The pre-synaptic variance $\vari{a}$ can enter the loss not at all or in one of two ways, respected by either the loss (see \citep{bayer2013training}) or the transfer function.
Three examples for this are 

\begin{enumerate}
    \item Squared loss on the mean, i.e. $d(y, z) = (\expc{y} - t)^2$ with $y = a$,
    \item Gaussian log-likelihood on the moments, i.e. $d(y, z) \propto {(\expc{y} - t) ^ 2 \over 2 \vari{y}} + \log \sqrt{2 \pi \vari{y}}$ with $y = a$,
    \item Negative Bernoulli cross entropy, i.e. $d(y, z) = z \log \expc{y} + (1 - z) \log (1 - \expc{y})$ with $y = {{1} \over {1 + \exp(-a)}}$.
\end{enumerate}

\begin{figure}
\includegraphics[width=.33 \textwidth]{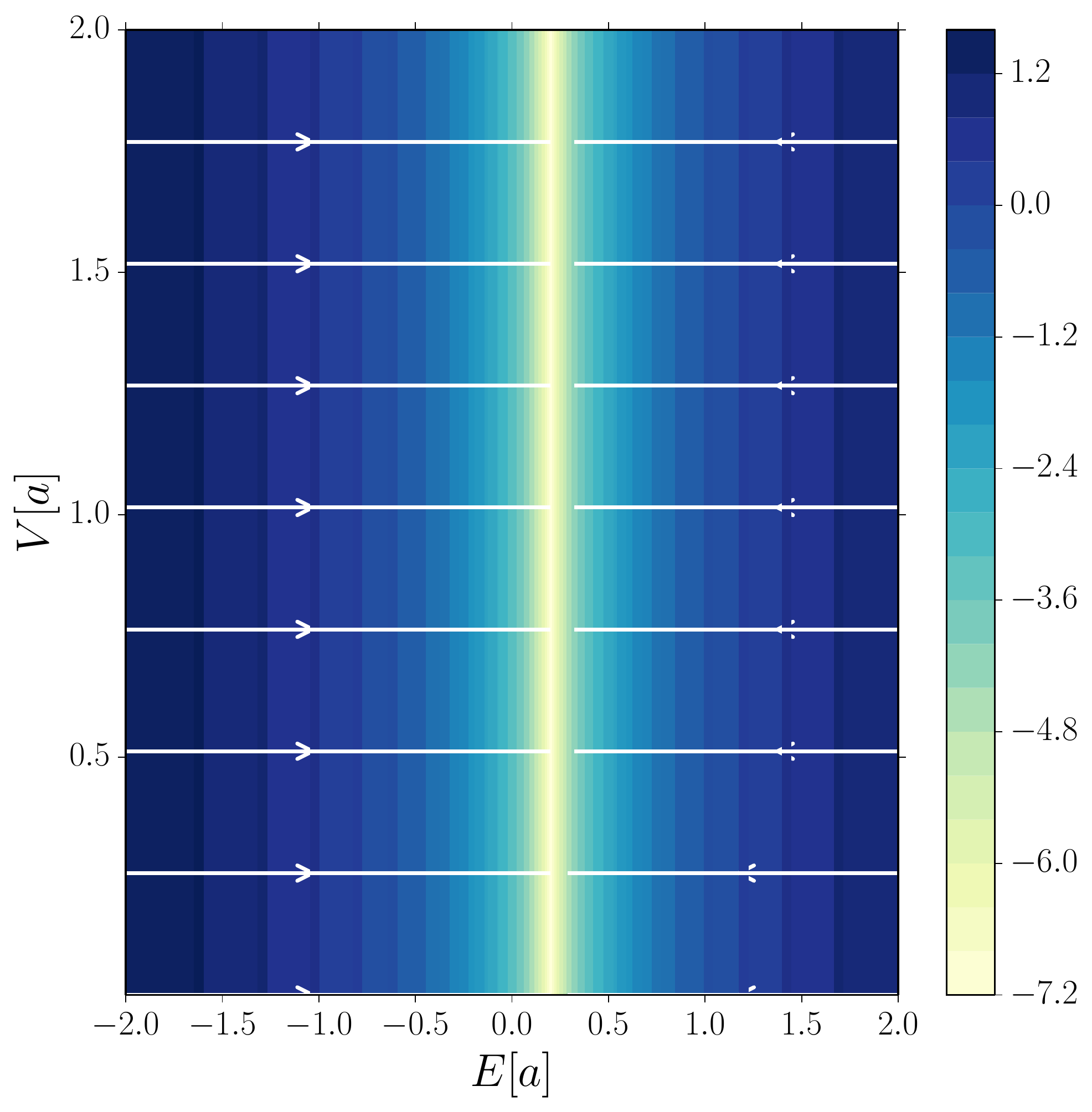}
\includegraphics[width=.33 \textwidth]{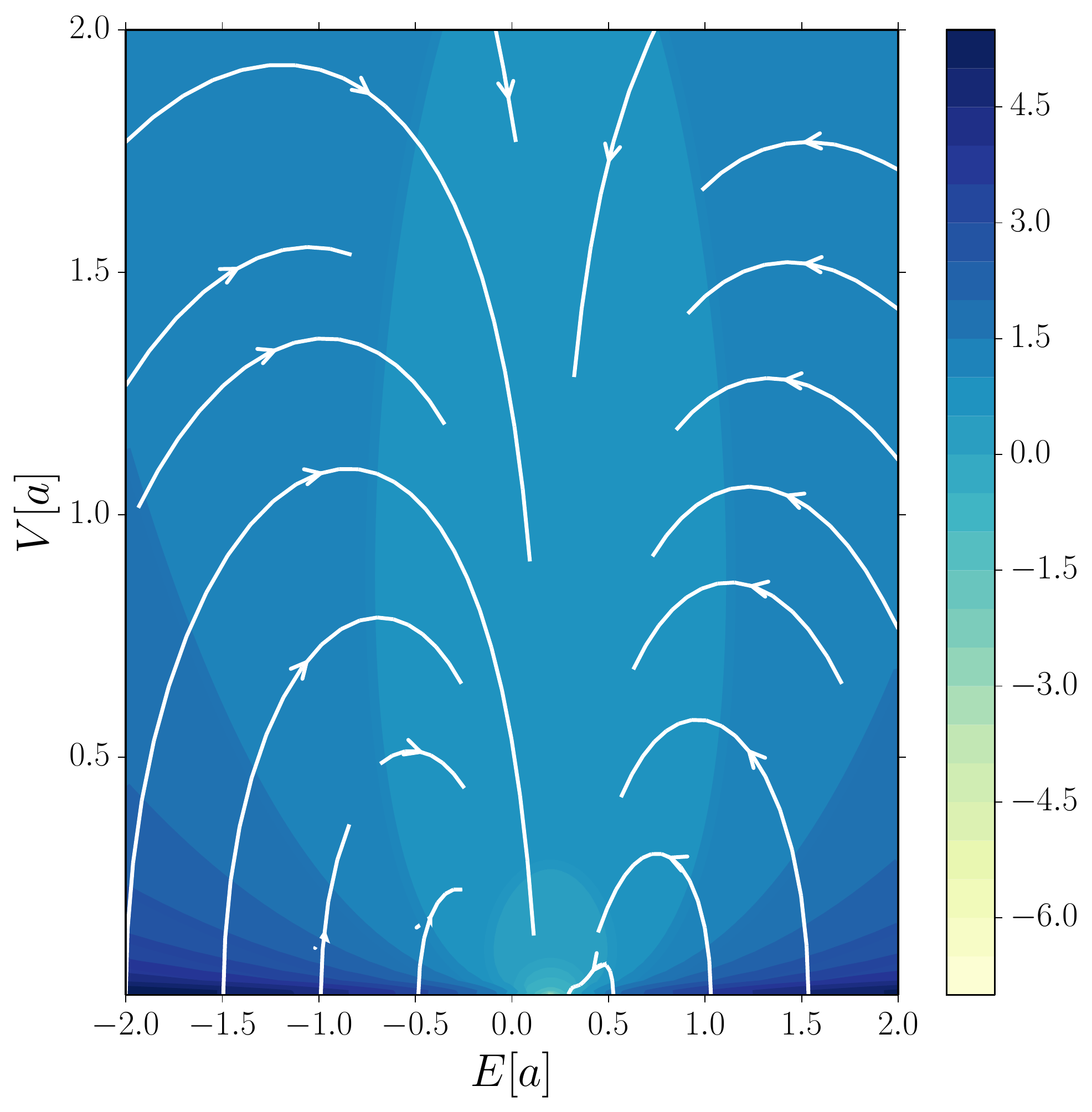}
\includegraphics[width=.33 \textwidth]{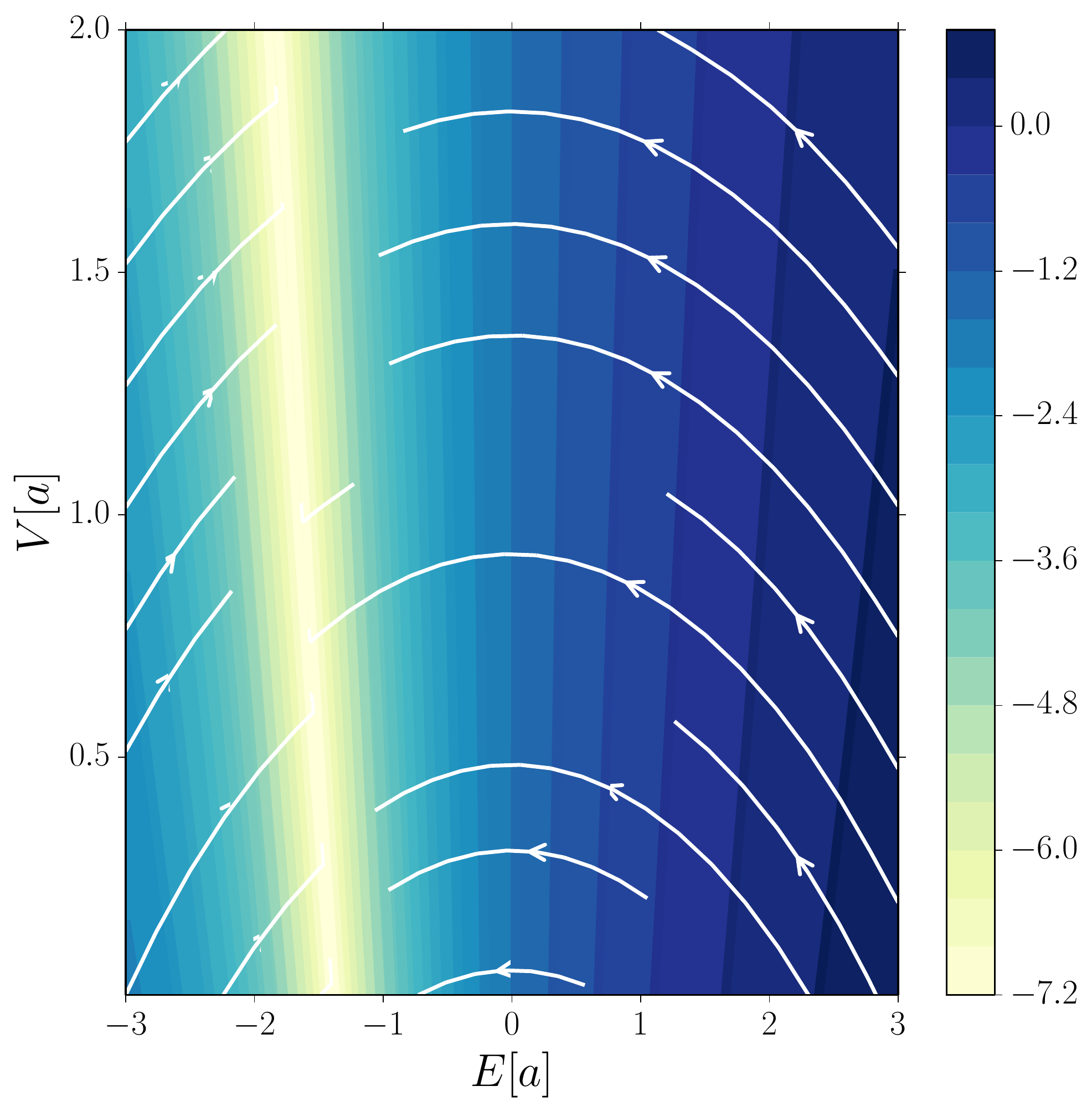}
\caption{
    Visualizations of the behaviour of $\delta^a$ for a single unit. 
    The axes correspond to the pre-synaptic mean $\expc{a}$ and variance $\vari{a}$ feeding into a unit $y = f(a)$. 
    A loss measuring the divergence from the target value $0.2$ is then applied and indices the color on a logarithmic scale.
    The gradients of the loss are shown as a vector field plot.
    Squared error is shown on the left, Gaussian log-likelihood in the middle and Bernoulli log-likelihood on the right.
    For the first two plots, the optimium is in the middle, for the last it is a little to the left.
}
\label{fig:delta}
\end{figure}

We visualize the pre-synaptic mean and variance, their gradients and their respective loss values in Figure~\ref{fig:delta}.
For the two latter cases, erroneous units first increase the variance, then move towards the correct mean and subsequently reduce the variance.

\paragraph{Sampling} An already mentioned alternative to the forward propagation of $\expc{a}$ and $\vari{a}$ through $f$ is to incarnate $\hat{a}$ via sampling and calculate $f(\hat{a})$.
Let $s \sim \mathcal{N}(0, 1)$ and $\hat{a} = \expc{a} + s  \sqrt{\vari{a}}$.
We can then use $\hat{a}$ explicitly and it follows that
\begin{eqnarray*}
{\partial \mathcal{J} \over \partial \hat{a}} {\partial \hat{a} \over \partial w_i} =
{\partial \mathcal{J} \over \partial \hat{a}} \bigg [
    {\partial \hat{a} \over \partial \expc{a}} {\partial \expc{a} \over \partial w_i}
    +
    {\partial \hat{a} \over \partial \vari{a}} {\partial \vari{a} \over \partial w_i}
\bigg ].
\end{eqnarray*}
Again, we recognize ${\partial \mathcal{J} / \partial \hat{a}} \cdot {\partial \hat{a} / \partial \expc{a}} \cdot {\partial \expc{a} / \partial{w_i}} = {\partial \mathcal{J} / \partial \hat{a}}~ x_i d_i$ as the standard back-propagation formula with dropout variables.

The variance term can be written as 
\begin{eqnarray*}
{\partial \mathcal{J} \over \partial \hat{a}}
{\partial \mathcal{J} \over \partial \vari{a}}
{\partial \vari{a} \over \partial w_i}
=
{\partial \mathcal{J} \over \partial \hat{a}}
{\partial \hat{a} \over \partial \sqrt{\vari{a}}}
{\partial \sqrt{\vari{a}} \over \partial \vari{a}}
{\partial \vari{a} \over \partial w_i^2}
{\partial w_i^2 \over \partial w_i}
\end{eqnarray*}
which, making use of results from earlier in the section is equivalent to
\begin{eqnarray}
{\partial \mathcal{J} \over \partial \hat{a}}
{\partial \mathcal{J} \over \partial \vari{a}}
{\partial \vari{a} \over \partial w_i}
&=&
{\partial \mathcal{J} \over \partial \hat{a}}
\sqrt{p(1-p) \expc{x_i}^2 + p\vari{x_i}} s.
\label{eq:gagaman}
\end{eqnarray}
The value of this is a zero-centred Gaussian random variable, since $s$ is Gaussian.
The scale is independent of the current weight value and only determined by the post-synaptic moments of the incoming unit, the dropout rate and the error signal.
We conclude, that also in this case, we can write
\begin{eqnarray*}
{\partial \mathcal{J} \over \partial w_i} &=&
    {\partial \mathcal{L}^a \over \partial w_i}
    + {\partial \mathcal{R}_{\text{draw}}^a \over \partial w_i},
\end{eqnarray*}
where ${\partial \mathcal{R}_{\text{draw}}^a / \partial w_i}$ is defined as in Equation (\ref{eq:gagaman}) and essentially an adaptive noise term.

We want to stress the fact that in the approximation as well as in the sampling case the regularization term vanishes at any optima of the training loss.
A consequence of this is that no global attractor is formed, which makes the method theoretically useful for RNNs. 
One might argue that fast dropout should not have a regularizing effect all.
Yet, regularization is not only influencing the final solution but also the optimization, leading to different optima.

\paragraph{Relationship to Weight Decay}
As already mentioned, imposing a Gaussian distribution centred at the origin with precision $\lambda$ as a prior on the weights leads to a method called weight decay.
It is not only probabilistically sound but also works well empirically, see e.g. \citep{bishop2006pattern}.
Recalling that the derivative of weight decay is of the form $\partial \mathcal{R}_{\text{wd}} / \partial w_i = 2 \lambda w_i$ we can reinterpret $\mathcal{R}_{\text{approx}}$ as a weight decay term where the coefficient $\lambda$ is weight-wise, dependent on the current activations and possibly negative.
Weight decay will always be slightly wrong on the training set, since the derivative of the weight decay term has to match the one of the unregularized loss.
In order for $\partial \mathcal{L} / \partial \theta + \partial \mathcal{R}_{\text{wd}} / \partial \theta$ to be minimal, $\mathcal{L}$ cannot be minimal unless so is $\mathcal{R}$.

\paragraph{Relationship to Adaptive Weight Noise}
In this method \citep{graves2011practical} not the units but the weights are stochastic, which is in pratice implemented by performing Monte Carlo sampling.
We can use a similar technique to FD to find a closed form approximation.
In this case, a layer is $y = f(\mathbf{x}^T\mathbf{w})$, where we have no dropout variables and the weights are Gaussian distributed with $\mathbf{w} \sim \mathcal{N}(\mu_{\mathbf{w}}, \sigma^2_{\mathbf{w}})$, with covariance diagonal and organized into a vector.
We assume Gaussian density for $a = \mathbf{x}^T\mathbf{w}$.
Using similar algebra as above, we find that 
\begin{eqnarray}
\label{eq:awn-exp}
\expc{a} & = & \mu_{\mathbf{x}}^T\mu_{\mathbf{w}}, \\
\vari{a} & = & \sigma_{\mathbf{x}}^{2T}\mu_{\mathbf{w}}^2 + \sigma_{\mathbf{x}}^{2T}\sigma_{\mathbf{w}}^2 + \mu_{\mathbf{x}}^{2T}\sigma_{\mathbf{w}}^2.
\label{eq:awn-var}
\end{eqnarray}
It is desirable to determine whether fast dropout and ``fast adaptive weight noise'' are special cases of each other.
Showing that the aproaches are different can be done by equating equations (\ref{eq:fd-exp}) and (\ref{eq:awn-exp}) and solving for $\mu_\mathbf{w}$.
This shows that rescaling by $1 - p$ suffices in the case of the expectation.
It is however not as simple as that for the variance, i.e. for equations (\ref{eq:fd-var}) and (\ref{eq:awn-var}), where the solution depends on $\mu_{\mathbf{x}}$ and $\sigma^2_{\mathbf{x}}$ and thus is not independent of the input to the network.
Yet, both methods share the property that no global attractor is present in the loss: the ``prior'' is part of the optimization and not fixed.

\subsection{Bag of Tricks}
Throughout the experiments we will resort to several ``tricks'' that have been introduced recently for more stable and efficient optimization of neural networks and RNNs especially.
First, we make use of \emph{rmsprop} \citep{tieleman2012rmsprop}, an optimizer which divides the gradient by an exponential moving average of its squares.
This approach is similar to Adagrad \citep{duchi2011adaptive}, which uses a window based average.
We found that enhancing \emph{rmsprop} with Nesterov's accelerated gradient \citep{sutskever2013training} greatly reduces the training time in preliminary experiments.

To initialize the RNNs to stable dynamics we followed the initialization protocol of \citep{sutskever2013importance} of setting the spectral radius $\rho$ to a specific value and the maximum amount of incoming connections of a unit to $\nu$; we did not find it necessary to centre the inputs and outputs.
The effect of not only using the recurrent weight matrix for propagating the states through time but also its element-wise square for advancing the variances can be quantified.
The stability of a network is coupled to the spectral radius of the recurrent weight matrix $\rho(\Wrec)$; thus, the stability of forward propagating the variance is related to the spectral radius of its element-wise square $\rho(\Wrec^2)$.
Since $\rho(AB) \le \rho(A)\rho(B)$ for non-negative matrices and non-singular matrices $A$ and $B$  \citep{horn2012matrix}, setting $\Wrec$ to full rank and its spectral radius to $\rho(\Wrec)$ assures $\rho(\Wrec^2) = \rho(|\Wrec|^2) \le \rho(\Wrec)^2$, where $|\cdot|$ denotes taking the absolute value element-wise.
We also use the gradient clipping method introduced in \citep{pascanu2012difficulty}, with a fixed threshold of 225.

Since the hidden-to-hidden connections and the hidden-to-output connections in an RNN can make use of hidden units in quite distinct ways, we found it beneficial to separate the dropout rates.
Specifically, a hidden unit may have a different probability to be dropped out when feeding into the hidden layer at the next time step than when feeding into the output layer.
Taking this one step further, we also consider networks in which we completely neglect fast dropout for the hidden-to-output connections; an ordinary forward pass is used instead.
Note that this is not the same as setting the dropout rate to zero, since the variance of the incoming units is completely neglected.
Whether this is done is treated as another hyperparameter for the experiment.

\section{Experiments and Results}

\begin{table}
    \caption[ ]{
        Results on the midi data sets. 
        All numbers are average negative log-likelihoods on the test set, where ``FD'' represents our work; ``plain'' and ``RNN-NADE'' results are from \citep{bengio2012advances} while ``Deep RNN`` shows the best results from \citep{pascanu2013construct}. 
        Note that ``RNN-NADE'' and ``Deep RNN`` employ various extensions of the model structure of this work, i.e. structured outputs and various forms of depths. 
        Our results are the best for the shallow RNN model considered in this work.
    }
    \label{table:results-midi}
    \begin{center}
    \begin{small}
    \begin{tabular}[t]{|l|r|r||r|r|}
        \hline
        Data set & FD & plain & RNN-NADE & Deep RNN \\
        \hline
        Piano-midi.de & 7.39 & 7.58 & 7.05 & -- \\
        Nottingham & 3.09 & 3.43 & 2.31 & 2.95  \\
        MuseData & 6.75 & 6.99 & 5.60 & 6.59 \\
        \belowspace
        JSBChorales & 8.01 & 8.58 & 5.19 & 7.92 \\
        \hline
    \end{tabular}
    \end{small}
    \end{center}
\end{table}


\subsection{Musical Data}
All experiments were done by performing a random search \citep{bergstra2012random} over the hyper parameters (see Table \ref{table:hyperpars} in the Appendix for an overview), where 32 runs were performed for each data set.
We report the test loss of the model with the lowest validation error over all training runs, using the same split as in \citep{bengio2012advances}.
To improve speed, we organize sequences into minibatches by first splitting all sequences of the training and validation set into chunks of length of 100.
Zeros are prepended to those sequences which have less than 100 time steps.
The test error is reported on the unsplitted sequences.

Training RNNs to generatively model polyphonic music is a valuable benchmark for RNNs due to its high dimensionality and the presense of long as well as short term dependencies.
This data set has been evaluated previously by \cite{bengio2012advances} where the model achieving the best results, RNN-NADE \citep{boulanger2013high}, makes specific assumptions about the data (i.e. binary observables).
RNNs do not attach any assumptions to the inputs.
\subsubsection{Setup}
The data consists of four distinct data sets, namely \emph{Piano-midi.de} (classical piano music), \emph{Nottingham} (folk music), \emph{MuseData} (orchestral) and \emph{JSBChorales} (chorales by Johann Sebastian Bach).
Each has a dimensionality of 88 per time step organized into different \emph{piano rolls} which are sequences of binary vectors; each component of these vectors indicates whether a note is occuring at the given time step.
We use the RNN's output to model the sufficient statistics of a Bernoulli random variable, i.e. 
\begin{eqnarray*}
p(x_{t ,i}|\mathbf{x}_{1:t-1}) = y_{t, i},
\end{eqnarray*}
which describes the probability that note $i$ is present at time step $t$.
The output non-linearity $f_y$ of the network is a sigmoid which projects the points to the interval $(0, 1)$.
We perform learning by the minimization of the average negative log-likelihood (NLL); in this case, this is the average binary cross-entropy
\begin{eqnarray*}
\mathcal{L}(\theta) = {1 \over T-1} {1 \over N} \sum_{i, t, k} x^{(k)}_{t, i} \log y^{(k)}_{t-1, i} + (1 - x^{(k)}_{t, i}) \log (1 - y^{(k)}_{t-1, i}),
\end{eqnarray*}
where $k$ indices the training sample, $i$ the component of the target and $t$ the time step.

\subsubsection{Results}
Although a common metric for evaluating the performance of such benchmarks is that of accuracy \citep{bay2009evaluation} we restrict ourselves to that of the NLL--the measure of accuracy is not what is optimized and to which the NLL is merely a proxy.
We present the results of FD-RNNs compared with the various other methods in Table \ref{table:results-midi}.
Our method is only surpassed by methods which either incorporate more specific assumptions of the data or employ various forms of depth \citep{boulanger2013high, pascanu2013construct}.
We want to stress that we performed only 32 runs for each data set once more.
This shows the relative ease to obtain good results despite of the huge space of potential hyper parameters.

\begin{figure}
\begin{center}
\includegraphics[width=.7 \textwidth]{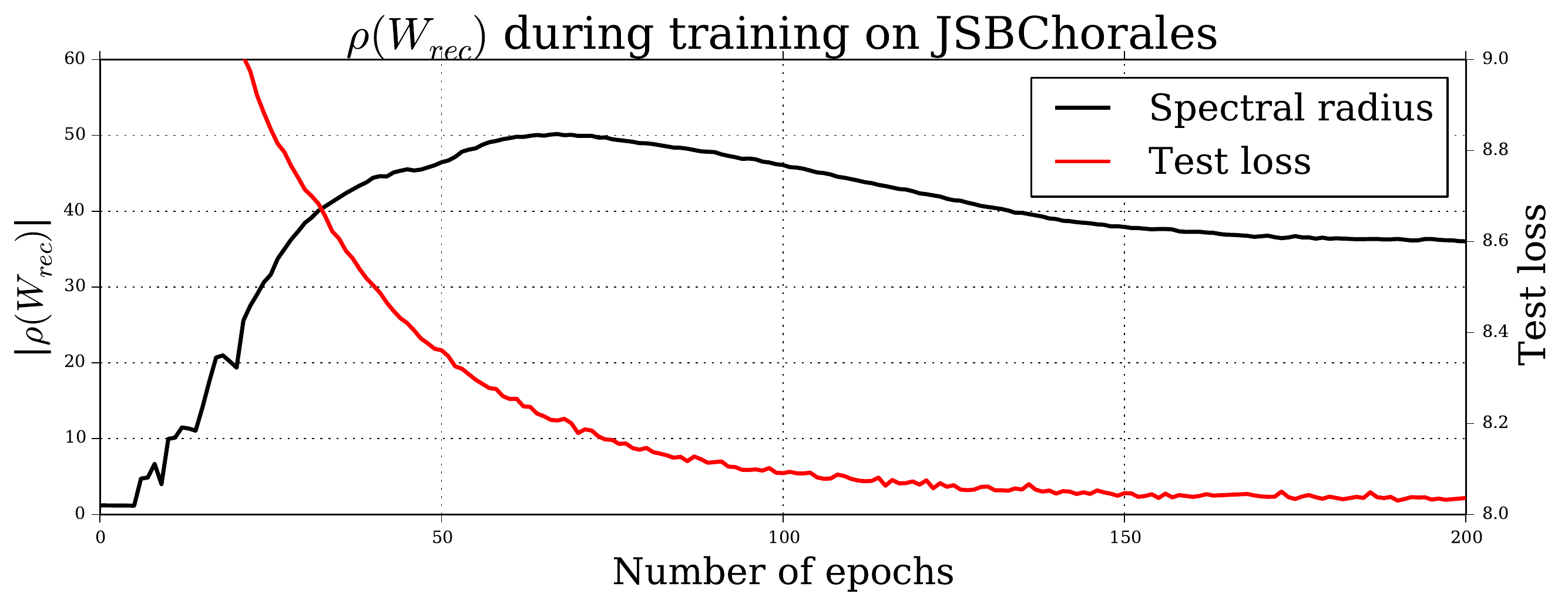}
\end{center}
\caption{
    Spectral radius over the training process.
    It increases first and then slowly decreases until a certain value is reached.
    We did not observe this behaviour when training plain RNNs.
}
\label{fig:eigen}
\end{figure}

One additional observation is the range of the Eigenvalues of the recurrent weight matrix $W_{\text{rec}}$ during training. 
We performed an additional experiment on \emph{JSBChorales} where we inspected the Eigenvalues and and the test loss.
We found that the spectral radius first increases sharply to a rather high value and then decreases slowly to settle to a specific value.
We tried to replicate this behaviour in plain RNNs, but found that RNNs never exceeded a certain spectral radius at which they stuck.
This stands in line with the observation from Section~\ref{subsubsec-bp} that weights are encouraged to grow when the error is high and shrink during convergence to the optimum.
See Figure~\ref{fig:eigen} for a plot of the spectral radius $\rho(\Wrec)$ over the training process stages.

\section{Conclusion}
We have contributed to the field of neural networks in two ways.
First, we have analysed a fast approximation of the dropout regularization method by bringing its derivative into the same form as that of a loss regularized with an additive term.
We have used this form to gain further insights upon the behaviour of fast dropout for neural networks in general and shown that this objective function does not bias the solutions to those which perform suboptimal on the unreguarized loss.
Second, we have hypothesized that this is beneficial especially for RNNs
We confirmed this hypothesis by conducting quantitative experiments on an already established benchmark used in the context of learning recurrent networks.


\bibliographystyle{apalike}
\bibliography{tex/bibliography}{}
\section{Appendix}

\subsection{Hyper Parameters for Musical Data Experiments}

We show the hyper parameters ranges for the musical data in Table~\ref{table:hyperpars}.
The ones from which the numbers in Table~\ref{table:results-midi} resulted are given in Table~\ref{table:hp-midi}.

\begin{table}
\caption[ ]{Hyper parameter ranges and parameter distributions for the musical data sets.}
\label{table:hyperpars}
\vskip 0.15in
\begin{center}
\begin{small}
\begin{tabular}[t]{|l|r|}
    \hline
    \abovespace\belowspace
    Hyper parameter & Choices\\
    \hline
    \abovespace
    \#hidden layers & $1$ \\
    \#hidden units & $200, 400, 600$ \\
    Transfer function & tanh \\
    $p(\text{``dropout input''})$ & $0.0, 0.1, 0.2$ \\
    $p(\text{``dropout hidden to hidden''})$ & $0.0, 0.1, 0.2, 0.3, 0.4, 0.5$ \\
    $p(\text{``dropout hidden to output''})$ & $0.0, 0.2, 0.5$ \\
    Use fast dropout for final layer & yes, no \\
    \abovespace
    Step rate & $0.01, 0.005, 0.001, 0.0005, 0.0001, 0.00001$\\
    Momentum & $0.0, 0.9, 0.95, 0.99, 0.995$ \\
    Decay & $0.8, 0.9$ \\
    \abovespace
    $W$ & $\mathcal{N}(0, \sigma^2), \sigma^2 \in \{0.1, 0.01, 0.001, 0.0001\}$ \\
    $\Win$ & $\mathcal{N}(0, \sigma^2), \sigma^2 \in \{0.1, 0.01, 0.001, 0.0001\}$ \\
    ${b_{h}}$ & $\mathbf{0}$ \\
    ${b_{y}}$ & $\mathbf{-0.8}$ \\
    \belowspace
    $\rho(\Wrec)$ & $1.0, 1.05, 1.1, 1.2$ \\
    $\nu$ & $15, 25, 35, 50$ or no \\
    \hline
\end{tabular}
\end{small}
\end{center}
\end{table}

\begin{table}[h]
\caption[ ]{Hyper parameters used for the musical data experiments.}
\label{table:hp-midi}
\vskip 0.15in
\begin{center}
\begin{small}
\begin{tabular}[h]{|l|r|r|r|r|}
    \hline
    \abovespace\belowspace
    Hyper parameter & Piano-midi.de & Nottingham & MuseData & JSBChorales \\
    \hline
    \abovespace
    \#hidden units & $600$ & $400$ & $600$ & $400$ \\
    $p(\text{``dropout input''})$ & $0.1$ & $0.1$ & $0.2$ & $0.1$  \\
    $p(\text{``dropout hidden to hidden''})$ & $0.3$ & $0.4$ & $0.3$ & $0.2$  \\
    $p(\text{``dropout hidden to output''})$ & -- & $0.0$ & $0.0$ & $0.5$ \\
    Use fast dropout for final layer & no & yes & yes & yes \\
    \abovespace
    Step rate & $0.005$ & $0.001$ & $0.0005$ & $0.001$ \\
    Momentum & $0.995$ & $0.99$ & $0.995$ & $0.99$ \\
    Decay & $0.8$ & $0.9$ & $0.9$ & $0.8$ \\
    \abovespace
    $\sigma^2$ for $\Wrec, \Wout$ & $0.1$ & $0.001$ & $0.1$ & $0.0001$ \\
    $\sigma^2$ for $\Win$ & $0.0001$ & $0.1$ & $0.0001$ & $0.01$ \\

    $\rho(\Wrec)$ & $1.2$ & $1.2$ & $1.2$ & $1.2$ \\
    \belowspace
    $\nu$ & $25$ & no & no & $15$ \\
    \hline
\end{tabular}
\end{small}
\end{center}
\end{table}

\end{document}